\documentclass[runningheads]{llncs}
\usepackage[T1]{fontenc}
\usepackage{graphicx}

\usepackage{todonotes}
\usepackage[hidelinks]{hyperref}
\usepackage{cleveref}
\usepackage{xcolor}
\usepackage{color} 
\usepackage{soul} 
\usepackage{multicol}
\usepackage{tabularx}
\usepackage{enumitem}
\usepackage{multirow}
\usepackage{csquotes}
\usepackage{xurl}

\begin{document}
\title{A Graph-based RAG for Energy Efficiency Question Answering}
%
%
\author{Riccardo Campi\inst{1} \and Nicolò Oreste Pinciroli Vago\inst{1} \and Mathyas Giudici\inst{1} \and Pablo Barrachina Rodriguez-Guisado\inst{2} \and Marco Brambilla\inst{1} \and Piero Fraternali\inst{1}}
\authorrunning{R. Campi et al.}
%
\institute{Politecnico di Milano, DEIB Department. Milano,  Italy\\
\email{\{firstname.lastname\} @polimi.it}
\and Voltiva Energy, Parque Científico de Murcia. Murcia, Spain \\
\email{pablo.barrachina@voltiva.energy}
}

\maketitle              

\begin{abstract}
In this work, we investigate the use of Large Language Models (LLMs) within a graph-based Retrieval Augmented Generation (RAG) architecture for Energy Efficiency (EE) Question Answering.
First, the system automatically extracts a Knowledge Graph (KG) from guidance and regulatory documents in the energy field. Then, the generated graph is navigated and reasoned upon to provide users with accurate answers in multiple languages. 
We implement a human-based validation using the RAGAs framework properties, a validation dataset comprising 101 question-answer pairs, and domain experts. Results confirm the potential of this architecture and identify its strengths and weaknesses. Validation results show how the system correctly answers in about three out of four of the cases ($75.2\pm2.7\%$), with higher results on questions related to more general EE answers (up to $81.0\pm4.1\%$), and featuring promising multilingual abilities ($4.4\%$ accuracy loss due to translation).

\keywords{Retrieval Augmented Generation (RAG) \and Knowledge Graph (KG) \and Large Language Model (LLM) \and Energy Efficiency \and Question Answering \and Multilingualism \and Sustainability.}
\end{abstract}

\section{Introduction}
%

The focus on Energy Efficiency (EE) is gaining importance in the energy sector, especially with the aim of reaching net zero emissions \cite{Lowitzsch2020}, as recently outlined by European Institutions \cite{Regulation-EU-2021-1119}.
Energy users 
are identified as key actors, especially when renewable energy sources are used \cite{Regulation-EU-2023-955}. The implementation by these key actors of the latest guidelines on energy savings \cite{Directive-EU-2023-1791,FOUITEH2023101582}, with the adoption of eco-friendly behaviors, is required to meet EE. Furthermore, EE is improved by matching energy consumption with the availability and production cycles of renewable energy sources 
\cite{Regulation-EU-2023-955}.

During the last few years, the increasing adoption of Large Language Models (LLMs) has offered opportunities to enhance the understanding and optimization of energy consumption to meet EE \cite{giudici2023assessing}. However, since these models may not fulfill the expectations when asked to provide factual answers, or when local regulations and socioeconomic context must be taken into account \cite{eichman2022reviewing,frieden2020collective}, Retrieval-Augmented Generation (RAG) systems address the matter by coupling the LLM with a knowledge base that formally describes domain-specific information \cite{ARSLAN20243781}.

Considering the above scenario, we propose a graph-based RAG architecture designed to offer users recommendations to help them achieve EE. Our solution automatically extracts knowledge from the source documents, containing domain-specific information about energy consumption, EE, regulations, and incentives. Then, it answers user queries by implementing a reasoning process that navigates the knowledge graph.  It handles multiple languages and extracts the semantics of the documents independently of their language. We assess the validity of the architecture through a human-based validation experiment using some key metrics proposed by the Retrieval Augmented Generation Assessment (RAGAs) framework \cite{es-etal-2024-ragas}, in collaboration with domain experts, and a validation dataset comprising 101 question-answer pairs.

\section{Background and Related Works}

This section presents relevant literature from both technical and energy sustainability perspectives, primarily focusing on Large Language Models (LLMs) and Retrieval Augmented Generation (RAG) systems.

\subsubsection{LLM.} In recent years, there has been significant growth in research on Large Language Models (LLMs), which are AI-based Natural Language Processing (NLP) models built on top of the Transformer architecture. These models exhibit proficiency in comprehending language and generating new content, with robust multilingual and summarization capabilities \cite{ARSLAN20243781}.
However, LLMs often provide nonsensical or entirely made-up answers when asked questions they do not know, such as domain-specific or vague ones \cite{giudici2023assessing,sanguinetti2024assessing}. These are known as hallucinations and are primarily caused by a lack of domain-specific knowledge or a failure to understand the context. Hallucinations can have significant negative effects because they are difficult to distinguish from accurate information, potentially spreading misinformation and eroding user trust \cite{hallucinations_legal_domain}.

\subsubsection{RAG.} To overcome these limitations, Retrieval-Augmented Generation (RAG) systems emerge as a powerful solution by combining the strengths of information retrieval with the generative capabilities of LLMs. This enables RAG systems to tackle knowledge-intensive tasks, such as handling domain-specific questions or citing the sources of retrieved information, resulting in more accurate and contextually relevant outputs \cite{ARSLAN20243781}.
While the simplest architectures leverage vector embeddings to store and retrieve their data \cite{RAG}, new graph-based architectures are emerging, allowing RAG to provide more accurate and relevant answers by relying on Knowledge Graphs (KGs) \cite{edge2024from}, especially for complex questions that require synthesizing information from multiple sources.

\subsubsection{Related work.} 
Arslan et al. \cite{ARSLAN2024114827} proposed the use of Energy Chatbot, a vector-based multi-source RAG with the aim of enhancing decision-making for Small and Medium-sized Enterprises by providing comprehensive Energy Sector insights through a Question Answering system. Key findings emphasize how the integration of a RAG significantly enhances the system's ability to deliver accurate, relevant, and consistent information, especially with the Llama3.1:8B model. However, a graph-based version of this prototype is still lacking.
Similarly, Bruzzone et al. \cite{bruzzone2023generative} coupled a RAG system to a GPT4-based chatbot to enhance urban planning simulations. The system dynamically simulates various urban scenarios, providing urban planners with accurate information and encouraging sustainable actions.
Another work \cite{fortuna2024naturallanguageinteractionhousehold} investigated graph-based RAG approaches to answer complex questions on electricity with the use of some publicly available electricity consumption KGs. Key findings reveal promising results in integrating RAG and LLMs with KGs for electricity-related topics. However, this approach does not include knowledge extraction from domain-specific documents nor their integration into KGs.
Lastly, in 2023, Giudici et al. \cite{giudici2023assessing} explored ways to enhance understanding and optimize energy consumption to meet EE using LLMs. Their chatbots responded fluently and coherently to general inquiries but fell short in accuracy when addressing domain-specific questions. 
However, no relevant literature exists at the moment of writing on improving EE for users and households using a graph-based RAG approach.

\section{Methodology}

\begin{figure}[t]
\centering
\includegraphics[width=0.9\columnwidth]{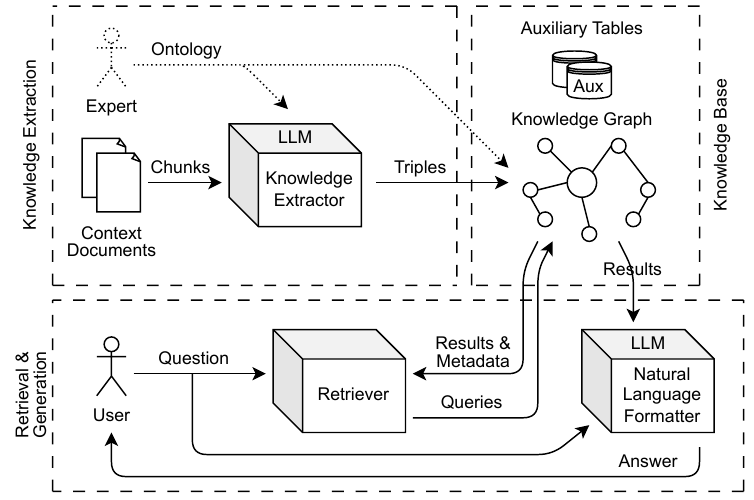}
\caption{General architecture of the proposed system. The system relies on 3 main parts: the \emph{Knowledge Extraction}, the \emph{Knowledge Base}, and the \emph{Retrieval \& Generation}.}
\label{fig:architecture}
\end{figure}

The proposed general architecture can be divided into 3 distinct parts, as shown in \Cref{fig:architecture}. First, a \emph{Knowledge Extractor} takes out triples containing entities and relationships from some domain-specific \emph{Context Documents}.  
To guide the extraction, a domain expert may inject their knowledge into the Extractor. The extracted triples are then analyzed and used to build the \emph{Knowledge Base}, containing a \emph{KG} and some auxiliary tables. Once the KG is populated, there is no need to rerun the extraction, unless it is necessary to add or remove domain-specific information. Finally, a \emph{Retrieval \& Generation} part receives the user's questions and uses a \emph{Retriever} to query the Knowledge Base to find relevant information. Then, a \emph{Natural Language Formatter} takes the question and the results from the interrogations and uses them to provide an answer to the user.

\subsection{Knowledge Extraction}

First, Context Documents (e.g., domain-specific documents, encompassing engineering and energy sector notions, available technologies, laws and regulations, and socioeconomic events) are provided to the system in the form of PDFs or web pages. These are cleaned of unnecessary parts, such as page numbers or HTML tags, and chunked into smaller chunks. The chunking algorithm divides text corpora into chunks using a chunk size and an overlapping size. When possible, the algorithm takes into consideration word boundaries, such as full stops or commas, to avoid splitting words within the same sentence between chunks.

The extraction phase then follows, consisting of the use of an LLM-based algorithm to parse the chunks with the aim of automatically extracting entities and relationships using a prompt-based approach. These are represented as entity-relationship-entity triples, where entities are usually objects and relationships are actions. When relevant, the algorithm adds properties to the nodes to enrich their semantics. Here, a domain expert may inject their knowledge to guide the system, focusing on specific aspects when extracting, eventually forcing it to adhere to a provided ontology or structure with a filtering algorithm. 

Finally, entity and relationship names are passed through a text-processing function that unifies their syntax. Similar names (e.g., \enquote{Energy Efficiency} and \enquote{energy\_efficiency}) are unified using the same syntax (e.g., \enquote{Energy efficiency}).

\subsection{Knowledge Base}
\label{subsec:knowledge_base}

It is designed to hold the KG with nodes and edges derived from the extraction, along with auxiliary tables to manage user metadata, such as their locations and preferences. It is necessary to populate the KG only once, before the system becomes fully operational or when it is time to update the information contained in it. The auxiliary tables can be filled instead at runtime based on user inputs.

First, the KG is initialized with a simple ontology that defines, under a new namespace ONTO, some objects of type \textit{OWL.Class} such as \textit{Entity}, \textit{Relationship}, \textit{Property}, \textit{Document}, \textit{Chunk}, etc. It also defines, under the same namespace, two new \textit{OWL.ObjectProperties} and \textit{OWL.DatatypeProperties} objects. Some examples of object properties are \textit{hasSource}, \textit{hasTarget}, \textit{hasRelationship}, and \textit{hasChunk}, while examples of datatype properties are \textit{hasName}, \textit{hasContent}, \textit{hasValue}, and \textit{hasValueEmbedding}.

The automatically extracted triples are then iteratively added to the KG as Entity and Relationship objects, as well as the source documents and their corresponding chunks, which become Document and Chunk objects, respectively. The nodes in the graph are identified by hashing the names of the objects. This allows the system to merge extracted entities into a single Entity if they represent the same thing, and the same applies to relationships and other objects. Edges between Entity and Relationship objects (e.g., \textit{hasRelationship}, \textit{hasSource}, \textit{hasTarget}, \textit{isSourceOf}, \textit{isTargetOf}, \textit{relatesTarget}, \textit{relatesSource}) are stored in such a way that allows for the precise reconstruction of the original extracted triple (i.e., the Entity-Relationship-Entity chain). To enable similarity-based local search and reasoning on graph nodes, embedding vectors are computed from Entity and Relationship names using a text embedding model, as well as from chunked text contents.

\subsection{Retrieval \& Generation}

Once the Knowledge Base is ready, the system allows users to ask questions and receive domain-specific answers.
The retrieval paradigm relies on a local, entity-based reasoning process. It first identifies relevant Entity objects in the KG by comparing them with the user question using a similarity measure, and then it begins a local reasoning process starting from the just-identified objects. Finally, it uses the information retrieved to provide an answer, augmenting it with citations to the original source documents. Answers are tailored for each specific user by using personalization metadata from the auxiliary tables.

When the Retriever receives a question, it extracts some triples using the same LLM-based algorithm as the extraction part. Then, it computes the embedding vectors of both the whole question and the extracted entity names, using the same text embedding model employed during graph construction. The just-mentioned vectors are then used in a similarity function (i.e., cosine similarity), which identifies the most similar Entity objects among the ones in the KG. The top-$k$ most similar ones are kept, with $3 \le k \le 15$ empirically selected.

If no entity exceeds the similarity threshold of $t$ (with $0.5 \le t \le 0.75$), the process moves to the Natural Language Formatter, which responds that no results exist for the current question. Otherwise, the process continues by identifying the top-$o$ outgoing and top-$i$ incoming Relationship objects to and from the Entity objects (with $5 \le o, i \le 10$ empirically selected). Also, the top-$c$ Chunk objects are retrieved ($5 \le c \le 10$). All these objects are then serialized as strings and used, along with other information such as the original question and the user metadata, to construct a prompt for the Natural Language Formatter, which in turn answers to the user.

\section{Validation experiment}

We conducted a validation experiment to assess the graph-based RAG architecture's ability to provide accurate, complete, and satisfactory answers in the EE domain. 
Tests are conducted using a dataset comprising EE question-answer pairs in various contexts and languages, based on guidance and regulatory documents from the EE field.


Implementation, deployment, and validation of our approach have been performed by incorporating our architecture into the \textit{ENERGENIUS Guru}, a Decision Support System (DSS) in the domain of \textit{EE} with a focus on the transparency and accountability of the knowledge sources, a contribution of the \textit{ENERGENIUS} European project\footnote{\url{https://energenius-project.eu/}}. Since the project is currently in its early stages, real usage data are still being collected. However, as a preliminary investigation, we created a dataset consisting of question-answer pairs that simulate real user questions.

\subsection{Validation dataset}

The dataset comprises a collection of source websites in Italian and a collection of 101 question-and-answer pairs about energy consumption, EE, regulations, and incentives extracted from the websites. The questions are divided as follows:

\begin{itemize}
    \item 25 questions \& answers focusing on Italian regulation and incentives on EE;
    \item 25 questions \& answers addressing Swiss regulations and incentives on EE;
    \item 51 questions \& answers providing recommendations and suggestions on EE, applicable to both Italy and Switzerland.
\end{itemize}

Questions and answers are made available in both Italian and English to assess the proposed system's ability to respond in a multilingual context. A complete list of source websites, in addition to some examples of question-answer pairs, is provided in \Cref{subsec:question_answer_pairs}.

\subsection{Experimental setup}

This section outlines the experimental setup utilized for conducting our tests.
Once the \emph{Context Documents} are downloaded from the just described source websites, we run the \emph{Knowledge Extraction} by cleaning text corpora from page numbers or HTML tags\footnote{Html2TextTransformer by LangChain (langchain\_community.document\_transformers)}, and then chunking them with $chunk\_size = 1000$, $chunk\_overlap = 200$\footnote{RecursiveCharacterTextSplitter by LangChain (langchain\_text\_splitters).}. Subsequently, we extract entities and relationships in the form of node-relationship-node triples using an LLM-based algorithm\footnote{LLMGraphTransformer by LangChain (langchain\_experimental.graph\_transformers)} using \emph{gpt-4o-mini} by OpenAI. For this test, we do not utilize domain-specific knowledge provided by a domain expert, since it is an optional feature.
Once extracted, we standardize the syntax of entity and relationship names by replacing all occurrences of \enquote{\texttt{\_}} with a space, converting the entire string to lowercase, and then capitalizing the first character only.
Below is provided a prompt example used to extract triples from text corpora: 

\scriptsize
\begin{verbatim}
You are a top-tier algorithm designed for extracting information
in structured formats to build a knowledge graph. Your task is to
identify the entities and relations requested with the user prompt
from a given text. You must generate the output in a JSON format
containing a list with JSON objects. Each object should have the
keys: "head", "head_type", "relation", "tail", and "tail_type".
[..]
The "relation" key must contain the type of relation between the
"head" and the "tail".
[..]
Attempt to extract as many entities and relations as you can.
Maintain Entity Consistency: When extracting entities, it's vital
to ensure consistency.
[..]
\end{verbatim}
\normalsize

We initialize a \emph{KG} in the \emph{Knowledge Base} using the simple ontology described in the \Cref{subsec:knowledge_base}. We then fill it by iteratively adding the just extracted Entity and Relationship objects. Objects in the KG are identified by hashing their names with MD5. We also compute their name embeddings using the \emph{text-embedding-3-small} model from OpenAI, as well as the chunked text corpora.

Once the Knowledge Base is ready, we simulate user interactions in \emph{Retrieval \& Generation} part by posing the previously mentioned questions from document sources to the system.
We pose questions both in Italian and English, and we ask the system to answer in the default user's language. This setting simulates the injection of user metadata from the \emph{Auxiliary Tables} to the Knowledge Base.
We used the same LLM model and embedding model as those used to populate the KG, \emph{gpt-4o-mini} and \emph{text-embedding-3-small} by OpenAI, respectively.
For each question, once obtained the most similar $k=12$ Entity objects (with $t=0.5$), their $c=5$ Chunk objects, and their $i=10$ ingoing and $o=10$ outgoing Relationship objects, the system provides these information to an LLM (in this case \emph{gpt-4o-mini} by OpenAI) that produces the final answer using this prompt (derived from a previous study \cite{edge2024from}): 

\scriptsize
\begin{verbatim}
---Role---
You are a helpful assistant responding to questions about data
in the tables provided.

---Goal---
Generate a response of the target length and format that responds
to the user's question, summarizing all information in the input
data tables appropriate for the response length and format, and
incorporating any relevant general knowledge.
If you don't know the answer, just say so. Do not make anything up.
Points supported by data should list their references as follows:
"This is an example sentence supported by multiple document
references [References: <page link>; <page link>]."

Do not list more than 5 record ids in a single reference. Instead,
list the top 5 most relevant record ids and add "+more" to
indicate that there are more.
[..]

Do not include information where the supporting evidence for it
is not provided.

---Target response length and format---
{"Single paragraph"}
Answer in {language}

---Data tables---
{context_data}

[..]

Add sections and commentary to the response as appropriate for
the length and format. Style the response in markdown.
\end{verbatim}
\normalsize

\section{Results}

The validation experiment takes the 101 questions in both Italian and English and produces as output a list of answers. As an example, \Cref{subsec:results} contains some of these answers. To assess whether our system is effective in answering EE-related questions, we adopt the human-based evaluation paradigm. In our case, this involves engaging domain experts, who are equipped with ground-truth answers, to assess and classify the responses as either valid or not by using the RAGAs framework guidelines \cite{es-etal-2024-ragas}. In particular, an answer is valid if it respects all three RAGAs metrics: \emph{faithfulness} (i.e., \enquote{the answer should be grounded in the given context}), \emph{answer relevance} (i.e., \enquote{the generated answer should address the actual question that was provided}), and \emph{context relevance} (i.e., \enquote{the retrieved context should be focused, containing as little irrelevant information as possible}). Experts must proportionally decrease an answer's score when it fails to meet one or more of the required properties until it reaches zero.

Based on our human-based evaluation conducted with $n=4$ domain experts, we achieve an overall answer validity score of $75.2\pm2.7\%$, with an average score of $77.4\pm2.9\%$ for responses in Italian language and $73.0\pm2.5\%$ for responses in English. A complete report, categorized by both question-answer language and country context, can be found in \Cref{tab:results}.

\begin{table}[ht]
    \renewcommand*{\arraystretch}{1.4}
    \centering
    \caption{Validation experiment results categorized by both language and context country. The experiment is conducted by asking domain experts to assess the answers using the RAGAs properties.}
    \label{tab:results}
    \begin{tabular}{p{2cm}|p{1.8cm}|p{1.8cm}|p{1.8cm}|p{1.8cm}|p{1.8cm}|}
        \multicolumn{2}{c}{\multirow{2}{*}{}} & \multicolumn{4}{c}{\textbf{Context country}:} \\
        \cline{3-6}
        \multicolumn{2}{c|}{} & \hfil \textbf{IT (25)} & \hfil \textbf{CH (25)} & \hfil \textbf{Both (51)} & \hfil \textbf{All (101)} \\
        \cline{2-6}
        \hfil\multirow{3}{*}{\textbf{Language}:} & \hfil \textbf{IT (101)} & \hfil $73.3\pm0.8\%$ & \hfil $74.4\pm1.9\%$ & \hfil $81.0\pm4.1\%$ & \hfil $77.4\pm2.9\%$ \\
        \cline{2-6}
         & \hfil \textbf{EN (101)} & \hfil $73.6\pm1.0\%$ & \hfil $67.7\pm2.4\%$ & \hfil $75.2\pm2.9\%$ & \hfil $73.0\pm2.5\%$ \\
        \cline{2-6}
         & \hfil \textbf{All (202)} & \hfil $73.4\pm0.9\%$ & \hfil $71.2\pm2.1\%$ & \hfil $78.1\pm3.0\%$ & \hfil $75.2\pm2.7\%$ \\
        \cline{2-6}
    \end{tabular}
\end{table}

Each question is answered in $19.08 \pm 4.48$ seconds on average, and involves a fixed number of 2 LLM calls and a variable number of embedding calls depending on the question’s complexity, averaging at $3.55 \pm 1.01$ calls per question.

\subsubsection{Ablation Experiment}

We performed an ablation experiment in which the RAG answered the 101 questions without any component representing persistent memory or retrieval. In this configuration, the system operates as an LLM-only architecture.
The obtained results show that retrieval technology is needed to answer domain-specific questions effectively. The system correctly answers basic or general questions on EE, but in most cases, answers are excessively long and sometimes contain inaccuracies. For instance, when asked why consumption is higher in winter, it replies that \enquote{winter causes more frequent baths or showers for hygiene reasons}. On the other hand, answers to specific questions, such as those about Italian or Swiss regulations, are mostly incorrect and often contain inaccuracies or irrelevant information. Sometimes, instead of providing a direct answer, they suggest the user search online for the information. For instance, this is the answer to the maximum deductible spending limit in Italy in 2025, which is set to 5,000 euros: \enquote{maximum deductible spending limit [..] in Italy for 2025 is set at 8,000 euros [..] please verify this information with the official sources}. 

\section{Discussion}

According to domain experts, the validation experiment results in \Cref{tab:results} indicate that the system produces valid responses in approximately three out of four of the cases.
When results are grouped by language, English responses perform almost as well as Italian ones, though slightly lower. This finding demonstrates how the multilingual capabilities of LLMs allow for valid semantic comprehension when information is provided or requested in different languages. The results show a $4.4\%$ reduction in answer accuracy due to translation errors. 
Finally, the ablation experiment demonstrated the need for systems tasked to answer domain-specific questions to have a retrieval component.

Categorizing results by context country shows how the system scores near the same for country-specific answers (i.e., answers for Italy and Switzerland), while it works slightly better for country-agnostic ones (i.e., answers marked as \enquote{both}). In our source documents, country-agnostic questions often convey general and discursive information, whereas country-specific ones refer more to specific laws and articles.
The slight decrease in performance for this class of questions could be due to laws or articles containing complex information linked by temporal or spatial constraints, which are harder to extract and manage compared to simpler information or general recommendations.

By categorizing the results both by context, country, and language, we confirm the findings highlighted above. In particular, performance is above average for country-agnostic concepts and in the Italian language, but slightly lower for country-specific concepts in languages other than the original documents.

\section{Conclusion}

Our research highlights the potential of an LLM-based system coupled with a KG-enhanced RAG architecture for EE.
This approach enables the system to offer tailored recommendations by integrating domain-specific knowledge, such as regulations and incentives on EE.

The LLM-based parsing process of documents enables the automated extraction of entity-relationship-entity triples, which typically does not require human intervention. If requested, a domain expert could impose the use of specific terminology or information on the extraction system.

The local reasoning process begins with the most relevant Entity objects in the KG and extends to their neighboring entities, ensuring that the answers are both accurate and contextually enriched. Additionally, the results show that performance in Italian and English is similar, indicating that LLMs can answer in languages other than the original source. This allows for a clear separation between the semantics of the content and its language.
Our validation experiment, which utilizes 101 question-answer pairs in two different languages and involves domain experts following the RAGAs framework guidelines, achieves an overall score of $75.2\pm2.7\%$, demonstrating the validity of our architecture in the field of EE. The best performance is achieved in country-agnostic questions in Italian, with $81.0\pm4.1\%$.

\subsubsection{Limitations.}
While this preliminary study highlights our architecture’s potential in aiding users with EE, as the \emph{ENERGENIUS} project advances, we will continuously gather and analyze data to enhance it and confirm these initial findings.
Moreover, testing the architecture should involve a variety of different LLMs and text embedding systems. To better validate multilingualism, it will be necessary to conduct tests in a broader range of languages beyond those already evaluated with source documents and question-answer pairs in this study.

\begin{credits}
\subsubsection{\ackname}
This research was funded by the European Union’s Horizon Europe Research and Innovation Framework, under Grant Agreement No 101160720. We acknowledge the use of AI-based tools for grammar corrections. 
We extend our gratitude to the domain experts, whose assistance was fundamental in conducting the validation experiment.

\end{credits}

%
%
%
\bibliographystyle{splncs04}
\bibliography{biblio}

\begin{thebibliography}{10}
\providecommand{\url}[1]{\texttt{#1}}
\providecommand{\urlprefix}{URL }
\providecommand{\doi}[1]{https://doi.org/#1}

\bibitem{ARSLAN20243781}
Arslan, M., Ghanem, H., Munawar, S., Cruz, C.: {A Survey on RAG with LLMs}. Procedia Computer Science  \textbf{246},  3781--3790 (2024), 28th Intl. Conf. on Knowledge Based and Intelligent information and Engineering Systems (KES 2024)

\bibitem{ARSLAN2024114827}
Arslan, M., Mahdjoubi, L., Munawar, S.: {Driving sustainable energy transitions with a multi-source RAG-LLM system}. Energy and Buildings  \textbf{324},  114827 (2024)

\bibitem{bruzzone2023generative}
Bruzzone, A., Giovannetti, A., Genta, G., Cefaliello, D.: {Generative AI and Retrieval-Augmented Generation (RAG) in an Agent-Based Simulation Framework for Urban Planning}. Int. Conf. on Modelling and Applied Simulation  (2023)

\bibitem{edge2024from}
Edge, D., Trinh, H., Cheng, N., Bradley, J., Chao, A., Mody, A., Truitt, S., Larson, J.: {From Local to Global: A Graph RAG Approach to Query-Focused Summarization} (2024), https://www.microsoft.com/en-us/research/publication/from-local-to-global-a-graph-rag-approach-to-query-focused-summarization/

\bibitem{eichman2022reviewing}
Eichman, J., Torrecillas~Castell{\'o}, M., Corchero, C.: {Reviewing and exploring the qualitative impacts that different market and regulatory measures can have on encouraging energy communities based on their organizational structure}. Energies  \textbf{15}(6), ~2016 (2022)

\bibitem{es-etal-2024-ragas}
Es, S., James, J., Espinosa~Anke, L., Schockaert, S.: {{RAGA}s: Automated Evaluation of Retrieval Augmented Generation}. In: Aletras, N., De~Clercq, O. (eds.) 18th Conf. of the European Chapter of the ACL: System Demonstrations. pp. 150--158. Association for Computational Linguistics, St. Julians, Malta (Mar 2024)

\bibitem{Directive-EU-2023-1791}
{European Parliament and Council of European Union}: {Directive (EU) 2023/1791 of the European Parliament and of the Council of 13 September 2023 on energy efficiency and amending Regulation (EU) 2023/955 (recast)} (2021)

\bibitem{Regulation-EU-2021-1119}
{European Parliament and Council of European Union}: {Regulation (EU) 2021/1119 of the European Parliament and of the Council of 30 June 2021 establishing the framework for achieving climate neutrality and amending Regulations (EC) No 401/2009 and (EU) 2018/1999 (''European Climate Law'')} (2021)

\bibitem{Regulation-EU-2023-955}
{European Parliament and Council of European Union}: {Regulation (EU) 2021/1119 of the European Parliament and of the Council of 10 May 2023 establishing a Social Climate Fund and amending Regulation (EU) 2021/1060} (2023)

\bibitem{fortuna2024naturallanguageinteractionhousehold}
Fortuna, C., Hanžel, V., Bertalanič, B.: {Natural Language Interaction with a Household Electricity Knowledge-based Digital Twin} (2024)

\bibitem{FOUITEH2023101582}
Fouiteh, I., {Cabrera Santelices}, J.D., Patel, M.K.: {How committed are swiss utilities to energy saving without being obligated to do so?} Utilities Policy  \textbf{82},  101582 (2023). \doi{10.1016/j.jup.2023.101582}

\bibitem{frieden2020collective}
Frieden, D., Tuerk, A., Neumann, C., d’Herbemont, S., Roberts, J.: {Collective self-consumption and energy communities: Trends and challenges in the transposition of the EU framework}. COMPILE, Graz, Austria  (2020)

\bibitem{giudici2023assessing}
Giudici, M., Abbo, G.A., Belotti, O., Braccini, A., Dubini, F., Izzo, R.A., Crovari, P., Garzotto, F.: {Assessing LLMs Responses in the Field of Domestic Sustainability: An Exploratory Study}. In: 2023 Third International Conference on Digital Data Processing (DDP). pp. 42--48. IEEE (2023)

\bibitem{RAG}
Lewis, P., Perez, E., Piktus, A., Petroni, F., Karpukhin, V., Goyal, N., K\"{u}ttler, H., Lewis, M., Yih, W.t., Rockt\"{a}schel, T., Riedel, S., Kiela, D.: {Retrieval-Augmented Generation for Knowledge-Intensive NLP Tasks}. In: Larochelle, H., Ranzato, M., Hadsell, R., Balcan, M., Lin, H. (eds.) Advances in Neural Information Processing Systems. vol.~33, pp. 9459--9474. Curran Associates, Inc. (2020)

\bibitem{Lowitzsch2020}
Lowitzsch, J., Hoicka, C.E., van Tulder, F.J.: {Renewable energy communities under the 2019 European Clean Energy Package – Governance model for the energy clusters of the future?} Renewable and Sustainable Energy Reviews  \textbf{122} (2020)

\bibitem{hallucinations_legal_domain}
Magesh, V., Surani, F., Dahl, M., Suzgun, M., Manning, C.D., Ho, D.E.: {Hallucination-Free? Assessing the Reliability of Leading AI Legal Research Tools}. Journal of Empirical Legal Studies  \textbf{22}(2),  216--242 (2025)

\bibitem{sanguinetti2024assessing}
Sanguinetti, M., Pani, A., Perniciano, A., Zedda, L., Loddo, A., Atzori, M.: {Assessing Italian Large Language Models on Energy Feedback Generation: A Human Evaluation Study}. 10th It. Conf. on Computational Linguistics (CLiC-it)  (2024)

\end{thebibliography}

\newpage
\appendix
\section{Appendix}

This section includes additional information about the validation experiment, specifically listing the source documents used to extract domain-specific knowledge and some of the question-answer pairs used to test the system.

\subsection{Source Documents}
\label{subsec:source_documents}

The source websites used to extract questions and answers for testing the system are as follows.

\footnotesize
\begin{itemize}
    \item Italian State Revenue Agency:
        \subitem \url{https://www.agenziaentrate.gov.it/portale/web/guest/aree-tematiche/casa/agevolazioni/bonus-mobili-ed-elettrodomestici}
    \item AEG Cooperativa:
        \subitem \url{https://www.aegcoop.it/lavatrice-risparmiare/}
        \subitem \url{https://www.aegcoop.it/migliori-lampadine/}
        \subitem \url{https://www.aegcoop.it/risparmiare-con-gli-elettrodomestici/}
        \subitem \url{https://www.aegcoop.it/consumi-standby-elettrodomestici/}
        \subitem \url{https://www.aegcoop.it/risparmiare-acqua-calda/}
        \subitem \url{https://www.aegcoop.it/riscaldamento-elettrico/}
    \item Luce-gas.it:
        \subitem \url{https://luce-gas.it/guida/risparmio-energetico}
    \item SvizzeraEnergia:
        \subitem \url{https://www.svizzeraenergia.ch/casa/} 
        \subitem \url{https://www.svizzeraenergia.ch/casa/riscaldamento/}
        \subitem \url{https://www.svizzeraenergia.ch/energie-rinnovabili/teleriscaldamento/}
    \item TicinoEnergia:
        \subitem \url{https://ticinoenergia.ch/it/domande-frequenti.html}
    \item Federal Department of the Environment, Transport, Energy and Communications:
        \subitem \url{https://www.uvek.admin.ch/uvek/it/home/datec/votazioni/votazione-sulla-legge-sull-energia/efficienza-energetica.html}.
\end{itemize}
\normalsize

\subsection{Question-answer pairs dataset}
\label{subsec:question_answer_pairs}

Here are some question-answer pairs from the dataset in both Italian and English. The system's responses, obtained from the validation experiment, are recorded in \Cref{subsec:results}.

\footnotesize
\begin{itemize}
    \item[IT] Qual è il tetto massimo di spesa detraibile nel 2025 per il bonus mobili ed elettrodomestici in Italia? --- 5.000 euro.
    \item[EN] What is the maximum deductible spending limit in 2025 for the furniture and household appliances bonus in Italy? --- 5.000 euros.
\end{itemize}

\begin{itemize}
    \item[IT] Quante ore di riscaldamento giornaliere nella zona climatica F in Italia? --- Nessuna limitazione.
    \item[EN] How many hours of heating per day in climate zone F in Italy? --- No limitations.
\end{itemize}

\begin{itemize}
    \item[IT] Quali sono i vantaggi delle lampadine LED? --- Minore consumo energetico, durata maggiore, eco-sostenibilità, illuminazione di elevata qualità e migliore tecnologia.
    \item[EN] What are the advantages of LED bulbs? --- Lower energy consumption, longer life, eco-sustainability, high-quality lighting and better technology.
\end{itemize}

\begin{itemize}
    \item[IT] Quanto consuma una lampadina LED? --- Tra 1 e 11 Watt all’ora.
    \item[EN] How much does an LED bulb consume? --- Between 1 and 11 Watts per hour.
\end{itemize}

\begin{itemize}
    \item[IT] Perché si consuma più acqua calda in inverno? --- A causa delle basse temperature.
    \item[EN] Why do we consume more hot water in winter? --- Due to the low temperatures.
\end{itemize}

\begin{itemize}
    \item[IT] Cosa è l'EtichettaEnergia in Svizzera? --- Un'etichetta che fornisce informazioni sugli apparecchi elettrici.
    \item[EN] What is the Energy label in Switzerland? --- A label providing information about electrical appliances.
\end{itemize}

\begin{itemize}
    \item[IT] Qual è la legge che definisce la superficie determinante nel Canton Ticino? --- La Legge edilizia cantonale (LE), art. 38, cpv. 3.
    \item[EN] What is the law that defines the determining surface area in the Canton of Ticino? --- The Cantonal Building Law (LE), art. 38, paragraph 3.
\end{itemize}
\normalsize

\subsection{Results}
\label{subsec:results}
This section presents the results of our validation experiment, including example answers to the questions from \Cref{subsec:question_answer_pairs}.

\footnotesize
\begin{itemize}
    \item[IT] Nel 2025, il tetto massimo di spesa detraibile per il bonus mobili ed elettrodomestici in Italia è fissato a 5.000 euro. Questa detrazione Irpef si applica per l'acquisto di mobili e grandi elettrodomestici destinati ad arredare un immobile oggetto di ristrutturazione. È importante notare che l'importo deve essere ripartito in dieci quote annuali di pari importo e il pagamento deve essere effettuato tramite bonifico o carta di debito/credito, escludendo altri mezzi di pagamento come assegni o contante [Ref: \url{https://www.agenziaentrate.gov.it/portale/web/guest/aree-tematiche/casa/agevolazioni/bonus-mobili-ed-elettrodomestici}].
    \item[EN] In 2025, the maximum deductible spending limit for the furniture and household appliances bonus in Italy is set at 5,000 euros. This amount includes any associated transportation and assembly costs, and the deduction must be divided into ten equal annual installments [Ref: \url{https://www.agenziaentrate.gov.it/portale/web/guest/aree-tematiche/casa/agevolazioni/bonus-mobili-ed-elettrodomestici}].
\end{itemize}

\begin{itemize}
    \item[IT] Nella zona climatica F, che è caratterizzata da gradi-giorno superiori a 3000, non è prevista alcuna limitazione per l'accensione degli impianti di riscaldamento. Ciò significa che il riscaldamento può essere attivato in qualsiasi momento dell'anno, senza restrizioni di giorni o orari [Ref: \url{https://luce-gas.it/guida/risparmio-energetico}].
    \item[EN] In climate zone F, there are no limitations on the heating period, meaning that heating can be utilized without specific restrictions on days or hours [Ref: \url{https://luce-gas.it/guida/risparmio-energetico}].
\end{itemize}

\begin{itemize}
    \item[IT] Le lampadine LED offrono numerosi vantaggi rispetto alle lampadine tradizionali. Innanzitutto, presentano un'efficienza energetica superiore, consentendo un risparmio di energia fino al 90\% rispetto alle lampadine alogene e a incandescenza, e fino all'80\% rispetto alle lampadine fluorescenti. Inoltre, la loro durata media è significativamente più lunga, variando tra 35.000 e 50.000 ore, rispetto alle sole 1.000 ore delle lampadine a incandescenza, riducendo così i costi di sostituzione. Infine, le lampadine LED non contengono sostanze chimiche tossiche, rendendole più ecologiche rispetto alle lampadine a risparmio energetico che possono contenere mercurio [Ref: \url{https://www.aegcoop.it/migliori-lampadine/}]. Questi fattori rendono le lampadine LED una scelta vantaggiosa sia dal punto di vista economico che ambientale.
    \item[EN] LED bulbs offer several significant advantages over traditional lighting options. Firstly, they are highly energy-efficient, consuming much less power; for instance, an LED bulb uses only 5 watts compared to 60 watts for an incandescent bulb, resulting in substantial energy savings. Additionally, LED bulbs have a longer lifespan, which reduces the frequency of replacements and contributes to lower overall costs. They also support advanced features such as dimmability and smart technology, allowing users to control brightness and color remotely. Furthermore, LED technology is more sustainable, making it a better choice for environmentally conscious consumers [Ref: \url{https://www.aegcoop.it/migliori-lampadine/}]. Overall, the combination of energy efficiency, longevity, versatility, and sustainability makes LED bulbs a superior lighting option.
\end{itemize}

\begin{itemize}
    \item[IT] Le lampadine LED hanno un consumo che varia generalmente tra i 3 e gli 11 Watt all'ora, con alcune tipologie che possono consumare solo 1 Watt all'ora. Questo è notevolmente ridotto rispetto alle lampadine tradizionali, poiché una lampadina LED da 11 Watt può emettere la stessa quantità di luce di una lampadina a incandescenza da 110-120 Watt. Ad esempio, una lampadina LED da 3 Watt consuma circa 0,003 kWh in un'ora di funzionamento, permettendo di illuminare un'intera casa con un consumo equivalente a quello di una singola lampadina a incandescenza [Ref: \url{https://www.aegcoop.it/migliori-lampadine/}].
    \item[EN] LED bulbs are known for their energy efficiency, typically consuming between **3 to 11 watts per hour**, with some models using as little as **1 watt per hour**. For instance, a **3-watt LED bulb** can produce the same amount of light as a traditional **110-120 watt incandescent bulb**. This significant reduction in power consumption translates to substantial savings on electricity bills, allowing an entire home to be illuminated with LED lights while consuming the same energy as a single incandescent bulb [Ref: \url{https://www.aegcoop.it/migliori-lampadine/}].
\end{itemize}

\begin{itemize}
    \item[IT] In inverno, il consumo di acqua calda tende ad aumentare a causa delle basse temperature esterne, che spingono le persone a cercare comfort attraverso l'uso di docce calde e bagni. Durante i mesi invernali, è comune consumare una quantità di acqua calda superiore rispetto al resto dell'anno, poiché le persone desiderano riscaldarsi e rilassarsi dopo una giornata fredda. Ad esempio, un bagno può richiedere fino a quattro volte più acqua rispetto a una doccia, contribuendo così a un aumento significativo della bolletta del gas o dell'elettricità, a seconda del tipo di riscaldamento utilizzato [Ref: \url{https://www.aegcoop.it/risparmiare-acqua-calda/}; \url{https://www.svizzeraenergia.ch/energie-rinnovabili/teleriscaldamento/}].
    \item[EN] In winter, the demand for hot water increases primarily due to the colder temperatures, which lead to higher usage for heating and personal hygiene. During this season, people tend to take longer showers and use more hot water for heating purposes, such as in radiators or underfloor heating systems. Additionally, appliances like dishwashers and washing machines often require hot water, further contributing to increased consumption. The relationship between winter months and hot water consumption is evident as colder weather necessitates more heating and comfort measures [Ref: \url{https://ticinoenergia.ch/it/domande-frequenti.html}; \url{https://www.svizzeraenergia.ch/casa/riscaldamento/}].
\end{itemize}

\begin{itemize}
    \item[IT] L'**etichettaEnergia** in Svizzera è un sistema di etichettatura che fornisce informazioni chiare e uniformi sui consumi energetici degli apparecchi elettrici, aiutando i consumatori a fare scelte più informate. Essa include dettagli come la classe di efficienza energetica, il consumo di energia, e altre caratteristiche importanti come il livello di rumore e il consumo d'acqua. Le etichette possono presentare un codice QR che consente l'accesso a ulteriori informazioni nella banca dati europea dei prodotti. L'etichettaEnergia si applica a vari gruppi di prodotti, tra cui frigoriferi e congelatori, e mira a proteggere sia l'ambiente che il portafoglio dei consumatori [Ref: \url{https://www.svizzeraenergia.ch/casa/}; \url{https://www.svizzeraenergia.ch/casa/riscaldamento/}; \url{https://www.svizzeraenergia.ch/energie-rinnovabili/teleriscaldamento/}].
    \item[EN] The Energy label in Switzerland, known as "Etichettaenergia," provides clear information about the energy consumption and efficiency of electrical appliances. It categorizes products on a scale from A to G, indicating their energy efficiency, and includes additional details such as noise levels and water consumption. The label is designed to help consumers make informed choices when purchasing appliances, promoting energy efficiency and environmental protection. In Switzerland, both QR-coded and non-QR-coded labels are accepted, ensuring that consumers can access further product information through a European database [Ref: \url{https://www.svizzeraenergia.ch/casa/riscaldamento/}; \url{https://www.svizzeraenergia.ch/casa/}].
\end{itemize}

\begin{itemize}
    \item[IT] La legge che definisce la superficie determinante nel Canton Ticino è l'articolo 38, comma 3 della Legge edilizia cantonale (LE). Secondo questa legge, la superficie determinante è parificata alla superficie edificata e si riferisce alla proiezione sul piano orizzontale dell’ingombro dell’edificio. Inoltre, è importante notare che l'impianto solare deve coprire una superficie pari al 50\% della superficie determinante [Ref: \url{https://ticinoenergia.ch/it/domande-frequenti.html}].
    \item[EN] The law that defines the determining surface area in the Canton of Ticino is outlined in the "Regolamento sull'utilizzazione dell'energia" (RUEn). This regulation specifies that the roofs and/or facades of new buildings with a determining surface area greater than 300 m² must be equipped with solar systems (both photovoltaic and thermal) until December 31, 2025 [Ref: \url{https://ticinoenergia.ch/it/domande-frequenti.html}].
\end{itemize}
\normalsize

\end{document}